\documentclass[letterpaper, 10 pt, journal, twoside]{ieeetran}

\usepackage{multirow}
\usepackage{graphicx} 

\usepackage{caption}
\usepackage{subcaption}

\usepackage{url}
\usepackage{xcolor}
\usepackage{algorithm}
\usepackage[noend]{algpseudocode} 

\usepackage{amssymb,amsmath,comment,cite}

\IEEEoverridecommandlockouts
\begin{document}

\title{MUI-TARE: Multi-Agent Cooperative Exploration with Unknown Initial Position}

\markboth{IEEE Robotics and Automation Letters. Preprint Version. June, 2022}
{Yan \MakeLowercase{\textit{et al.}}: MUI-TARE: Multi-Agent Cooperative Exploration with Unknown Initial Position}

\author{Jingtian Yan$^{1}$, 
        Xingqiao Lin$^{1}$, 
        Zhongqiang Ren$^{1}$,
        Shiqi Zhao$^{2}$,
        Jieqiong Yu$^{1}$, 
        Chao Cao$^{1}$,\\
        Peng Yin$^{1,\mathbf{*}}$, 
        Ji Zhang$^{1}$ and Sebastian Scherer$^{1}$

\thanks{
This research was supported by grants from NVIDIA and utilized NVIDIA SDKs (CUDA Toolkit, TensorRT, and Omniverse).

$^{1}$J. Yan, X. Lin, P. Yin, and S. Scherer are with Robotics Institute, Carnegie Mellon University, Pittsburgh, PA 15213, USA {(pyin2, basti@andrew.cmu.edu)}.}
}

\maketitle

\begin{abstract}


Multi-agent exploration of a bounded 3D environment with unknown initial positions of agents is a challenging problem.
It requires quickly exploring the environments as well as robustly merging the sub-maps built by the agents.
We take the view that the existing approaches are either aggressive or conservative:
Aggressive strategies merge two sub-maps built by different agents together when overlap is detected, which can lead to incorrect merging due to the false-positive detection of the overlap and is thus not robust.
Conservative strategies direct one agent to revisit an excessive amount of the historical trajectory of another agent for verification before merging, which can lower the exploration efficiency due to the repeated exploration of the same space.
To intelligently balance the robustness of sub-map merging and exploration efficiency, we develop a new approach for lidar-based multi-agent exploration, which can direct one agent to repeat another agent's trajectory in an \emph{adaptive} manner based on the quality indicator of the sub-map merging process.
Additionally, our approach extends the recent single-agent hierarchical exploration strategy to multiple agents in a \emph{cooperative} manner by planning for agents with merged sub-maps together to further improve exploration efficiency.
Our experiments show that our approach is up to 50\% more efficient than the baselines on average while merging sub-maps robustly. 
\end{abstract}


\section{Introduction}

    \begin{figure}[!t]
    \centering
    \includegraphics[width=0.98\linewidth]{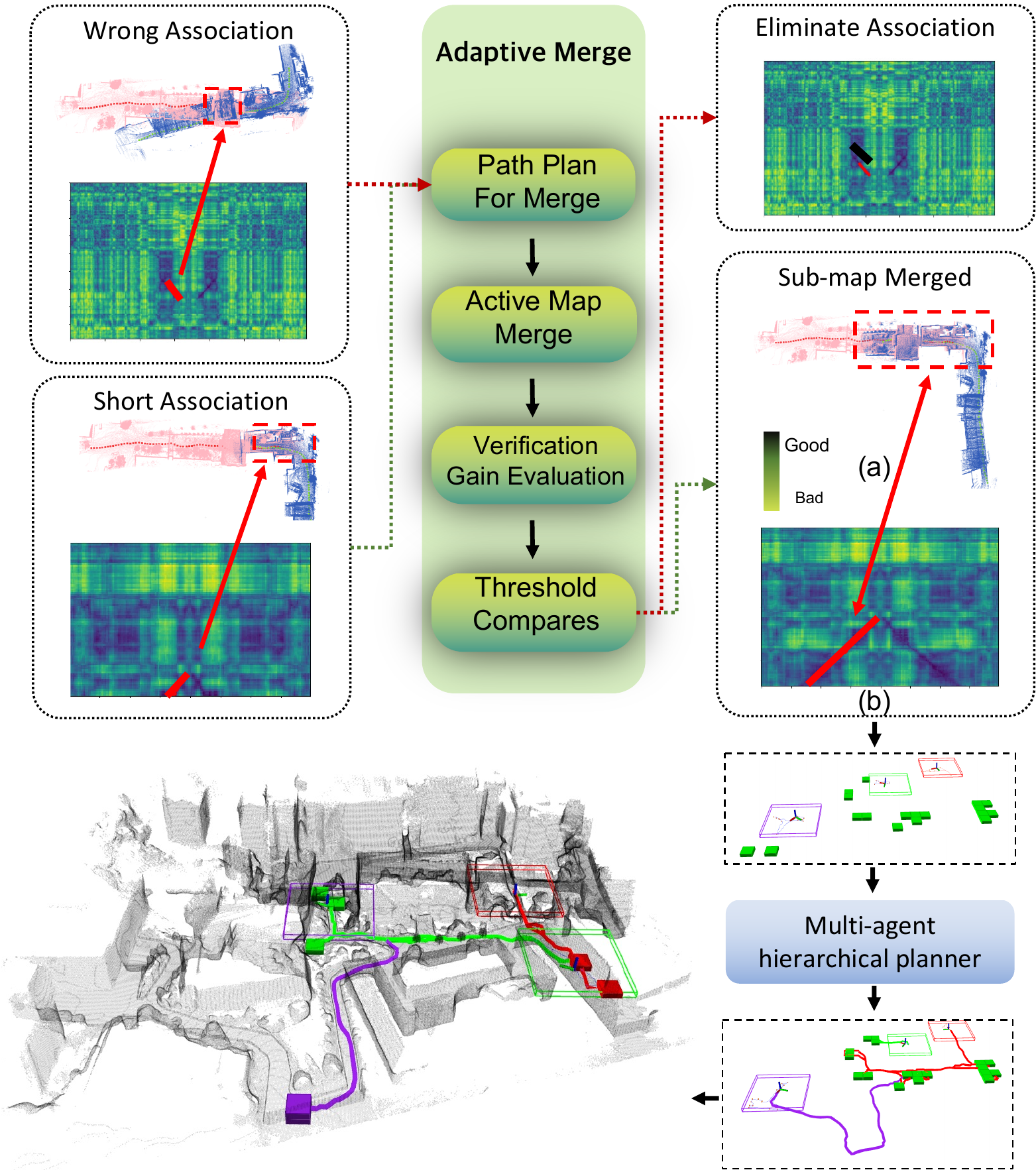}
    \caption{
        \textbf{Multi-agent exploration using Adaptive Merge.}
        (a) represents the point clouds merged using estimate transform, and (b) represents the data association result between the features of two agents (i.e. each axis represents the trajectory of one agent). Once a sequence of features gets matched, the adaptive merge module will verify the association by measuring the verification gain. 
        If the data association is incorrect, the association will be eliminated. Otherwise, it will adaptively increase the overlap and merge it into a sub-map.
        Using the relative pose and merged map given by the merged sub-map, the multi-agent sub-map exploration module is then used for exploration path planning.
    }
    \label{fig:idea}
    \end{figure}
    
    \IEEEPARstart{M}{ulti-Agent} exploration of an unknown environment is an important problem in robotics and has been investigated for decades~\cite{1435481, verma2021multi, 9492802}.
    In this work, we consider the exploration of a bounded 3D environment with unknown initial positions of agents, which arises in applications such as planetary exploration~\cite{schuster2019towards, hassanalian2018evolution}, underground mining~\cite{8741532} and search and rescue~\cite{999224, shah2004survey}.
    This problem is challenging as it requires simultaneously handling two goals: (i) quickly exploring the environment, and (ii) robustly detecting the overlapped sub-maps built by different agents and merging the sub-maps.
    A common strategy to address the problem is to directly merge the sub-maps when the areas explored by different agents are detected to overlap. 
    However, the sub-map merging procedure requires extracting features from each sub-map and associating the features across the sub-maps before merging them, which is often sensitive to false-positive feature matching~\cite{yu2020review}. In other words, a few mismatched features can lead to incorrect sub-map merging.
    To improve the robustness against false-positive feature matching, the idea of active verification has been applied~\cite{8967932}: when two agents are detected to have visited the same place, one agent is directed to repeat a certain amount of the other agent's historical trajectory in order to increase the overlapped region and improve the robustness of feature association. However, this strategy achieves robustness at the cost of exploration efficiency, as it often leads to excessive duplicated exploration among agents than necessary.

    In this paper, we propose MUI-TARE ({\textbf{\underline M}}ulti-agent {\textbf{\underline {TARE}}} with {\textbf{\underline U}}nknown {\textbf{\underline I}}nitial position), a multi-agent lidar-based exploration system that aims to intelligently balance between sub-map merging robustness and exploration efficiency.
    As shown in Fig. \ref{fig:idea}, the system has two features: adaptive sub-map merging and cooperative multi-agent exploration.
    First, MUI-TARE leverages the idea of active verification but in an adaptive manner:
    By leveraging our prior work AutoMerge~\cite{yin2022automerge}, a sub-map merging method, MUI-TARE is able to learn from AutoMerge the quality of the current feature association and then increases the overlapped region only when needed so that a certain quality threshold on feature matching is reached in AutoMerge.
    This allows an agent to repeat another agent's trajectory in an adaptive manner (based on the quality score) and thus avoid a large amount of duplicated exploration which leads to improvement in the overall exploration efficiency.
    Second, MUI-TARE employs a hierarchical cooperative multi-agent planning strategy for exploration, which extends a recent state-of-the-art single-agent exploration planner TARE~\cite{cao2021tare} to multiple agents.
    The hierarchical strategy first plans a coarse global path that visits the areas to be explored based on the global information of the agent's map and then plans a detailed local path to explore the current local area around the robot before moving to the next local area.
    Instead of naively applying this strategy to multiple agents, our approach plans for agents cooperatively when possible: it begins by planning each agent independently (before any sub-map is merged) and plans for agents together when their sub-maps are merged.
    Our approach improves the exploration efficiency due to this cooperative planning strategy.
    
    To verify the idea of adaptive verification and cooperative multi-agent planning, we run simulations in several environments. Our numerical results show that (i) our adaptive verification strategy leads to up to 29\% higher exploration efficiency than the existing active verification strategy while maintaining the robustness in the sub-map merging process; and (ii) the cooperative multi-agent planning reduces the overall exploration time for up to 52\% than naively applying TARE to multiple agents without cooperation.
    
    
	
	The rest of this work reviews the related work in Sec.~\ref{sec:related} and describes the problem in Sec.~\ref{sec:preli} We then describe our method in Sec.~\ref{sec:method} and show the simulation results in Sec. \ref{sec:experiments}. Finally, we conclude and outline our future work in Sec. \ref{sec:future-work}.

\section{Related Work}\label{sec:related}

    \subsection{Autonomous Exploration}
    Autonomous exploration using either single or multiple agents has been investigated for decades from different perspectives. 
    Single-agent exploration has been extensively studied in both 2D and 3D, and various approaches have been developed ranging from frontier-based~\cite{613851,9387089}, receding horizon sampling-based~\cite{7487281}, information theoretic approaches~\cite{1041446, stachniss2005information, carrillo2018autonomous}, traveling salesman-based~\cite{cao2021tare}, etc.
    Our approach is an extension of the recent hierarchical traveling salesman-based approach TARE~\cite{cao2021tare} to multiple agents.
    
    Multi-agent exploration is more challenging than single-agent and existing work has focused on communication limitation~\cite{amigoni2017multirobot}, sub-map merging when initial positions of agents are unknown~\cite{8967932}, allocating the unexplored areas to agents in a cooperative manner~\cite{1435481}, etc.
    In this work, we limit our focus to the problem where agents' initial locations are unknown.
    Many existing multi-agent exploration works are limited to 2D environment~\cite{1435481,fox2006distributed,8967932,yu2021smmr}, and extending them to 3D is challenging due to the increased amount of sensory data (e.g. 3D lidar scans), a large number of frontiers (i.e., the boundary between the explored and unexplored areas), the increased risk of feature mismatch when merging the sub-maps, etc.
    Inspired by the literature, we compare our approach against several baselines that take similar strategies to merge sub-maps as in~\cite{8967932,yu2021smmr} to verify the effectiveness of our approach.

    \begin{figure*}[!t]
        \centering
        \includegraphics[width=\linewidth]{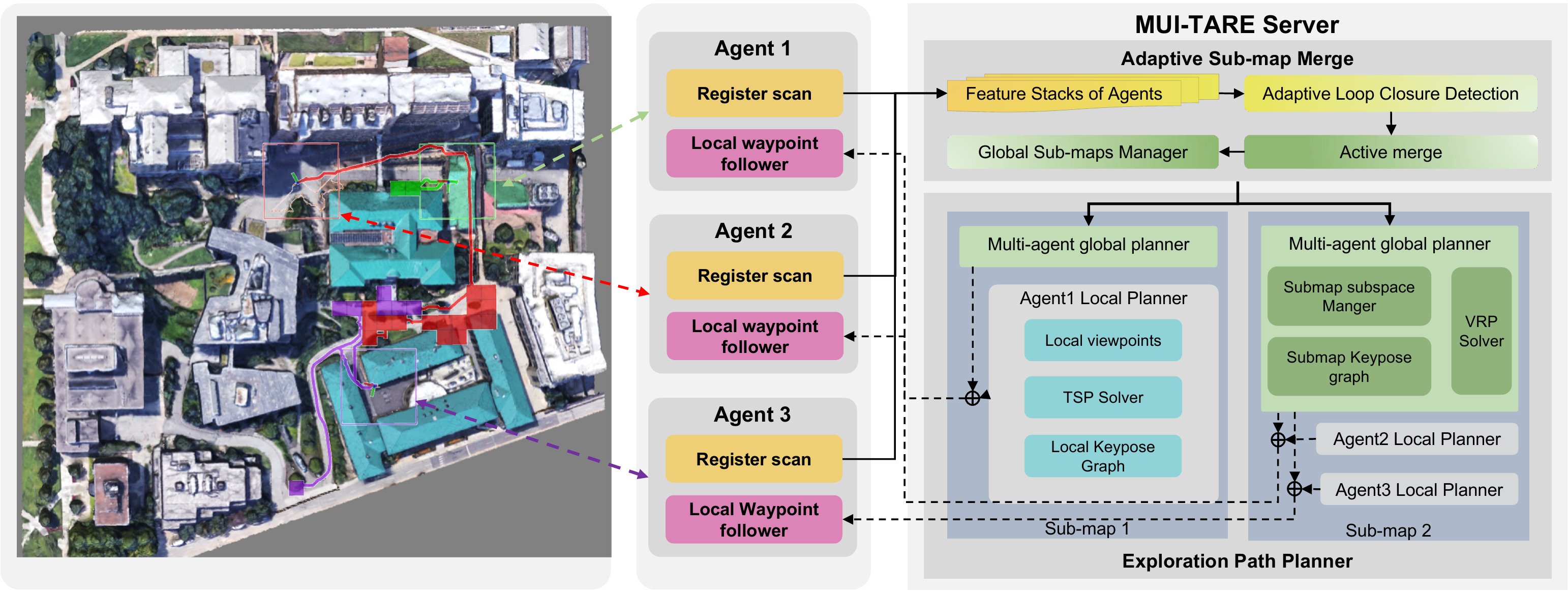}
        \caption{\textbf{The overall framework of MUI-TARE.}
        The overall system comprises two components: the adaptive sub-map merge and sub-map multi-agent exploration.
        In the adaptive map merge module, the inner connections of the agents are increased by actively exploring the overlapped region. The merged sub-map is provided along with the relative pose of agents on the sub-map.
        Given the sub-map, the MUI-TARE uses a hierarchical planner for the exploration path planning of each agent.}
        \label{fig:OverallFramework}
    \end{figure*}
    
    \subsection{Sub-map merging and data association}
    Map merging problems in 3D require integrating different sub-maps together, and the map is usually represented as 3D point clouds, occupancy grids \cite{carpin2008fast, hornung2013octomap}, or 3D meshes \cite{rosinol2021kimera}.
    In this work, we limit our focus to lidar-based multi-agent exploration and consider maps represented as point clouds.
    To merge two sub-maps accurately and robustly, a key step is to identify unique features from each of the maps and associate the detected features.
    The SLAM community have proposed numerous methods by leveraging distinguishable semantic objects \cite{dube2020segmap, yin2022automerge}, bag-of-words vectors~\cite{6202705}, learned full-image descriptors~\cite{arandjelovic2013all}, and sequence-based place recognition~\cite{6224623, 9341727}, to name a few.
    Our prior work AutoMerge~\cite{yin2022automerge} provides a large-scale map alignment approach for multi-agent systems by leveraging a novel spherical harmonic feature extraction method for viewpoint-invariant place recognition and an adaptive sequence alignment method for accurate loop-closure detection. 
    This work leverages AutoMerge to merge sub-maps and estimate the relative poses between agents.
    
    In most SLAM systems, the agents are passive, i.e., the motion of the agents cannot be controlled.
    In a multi-agent exploration problem, the motion of agents needs to be planned in order to achieve both high exploration efficiency and high quality in sub-map merging, which is the focus of this work.

\section{Problem Statement}\label{sec:preli}
Let ${Q \subset \mathbb{R}^3}$ denote a bounded space to be explored and let ${Q_{trav}} \subseteq {Q}$ denote the traversable space in ${Q}$.
Additionally, let ${S} \subset {Q}$ denote all the surfaces in the ${Q}$, which represents the generalized boundary between obstacle and obstacle-free space.
Let ${I}=\{1,2,\dots,N\}$ represent a set of $N$ agents with unknown initial poses in ${Q_{trav}}$.
Agents are allowed to exchange information with a central station at any time.
Let $ u_i^t = \{p_{u_i^t}, q_{u_i^t}\}, p_{u_i^t} \in Q_{trav}$ denote a viewpoint of the sensor onboard agent ${i} \in {I}$ at time ${t}$ with ${p_{u_i^t}}$ and ${q_{u_i^t}}$ denoting the position and orientation of the sensor respectively.
For each viewpoint, the surface perceived is represented as ${S_{v_i^t}^{cov}}$. 
Let ${\tau_{i}}$ denote a path of agent ${i \in I}$, and let $dist({\tau_{i}})$ represent the distance traveled by the agent along the path.
A sequence of viewpoints ${U^i} = \{{{u_i^0}, {u_i^1}, \dots, {u_1^{T_0}}}\} \subset {Q_{trav}}$ is visited along the path, and the path must satisfy the kinodynamic constraints of the agent.
The goal of the problem is to plan a set of kinodynamically feasible paths $ \tau = \{\tau_{1},\tau_{2},\dots, \tau_{N}\}$ such that the perceived surface of all agents' viewpoints along the paths covers ${S}$ (i.e., $\bigcup_1^N \bigcup_0^{T_i} {S_{v_i^t}^{cov}} = {S}$) and 
the longest travel distance of all the paths (i.e., $\max_{i\in I} dist(\tau_{i})$) is minimized.

\section{Method}
\label{sec:method}
This section begins with a system overview of MUI-TARE in Sec.~\ref{sec:system_overview}.
Instead of merging the point cloud from the agents passively, the framework used an adaptive sub-map merge algorithm to merge the sub-maps robustly, which is discussed in Sec.~\ref{sec:adaptive_merge}. Additionally, MUI-TARE utilizes a hierarchical multi-agent exploration method to efficiently explore each sub-map as presented in Sec.~\ref{sec:explore_submap}.

    \begin{figure}[!t]
    \centering
        \includegraphics[width=\linewidth]{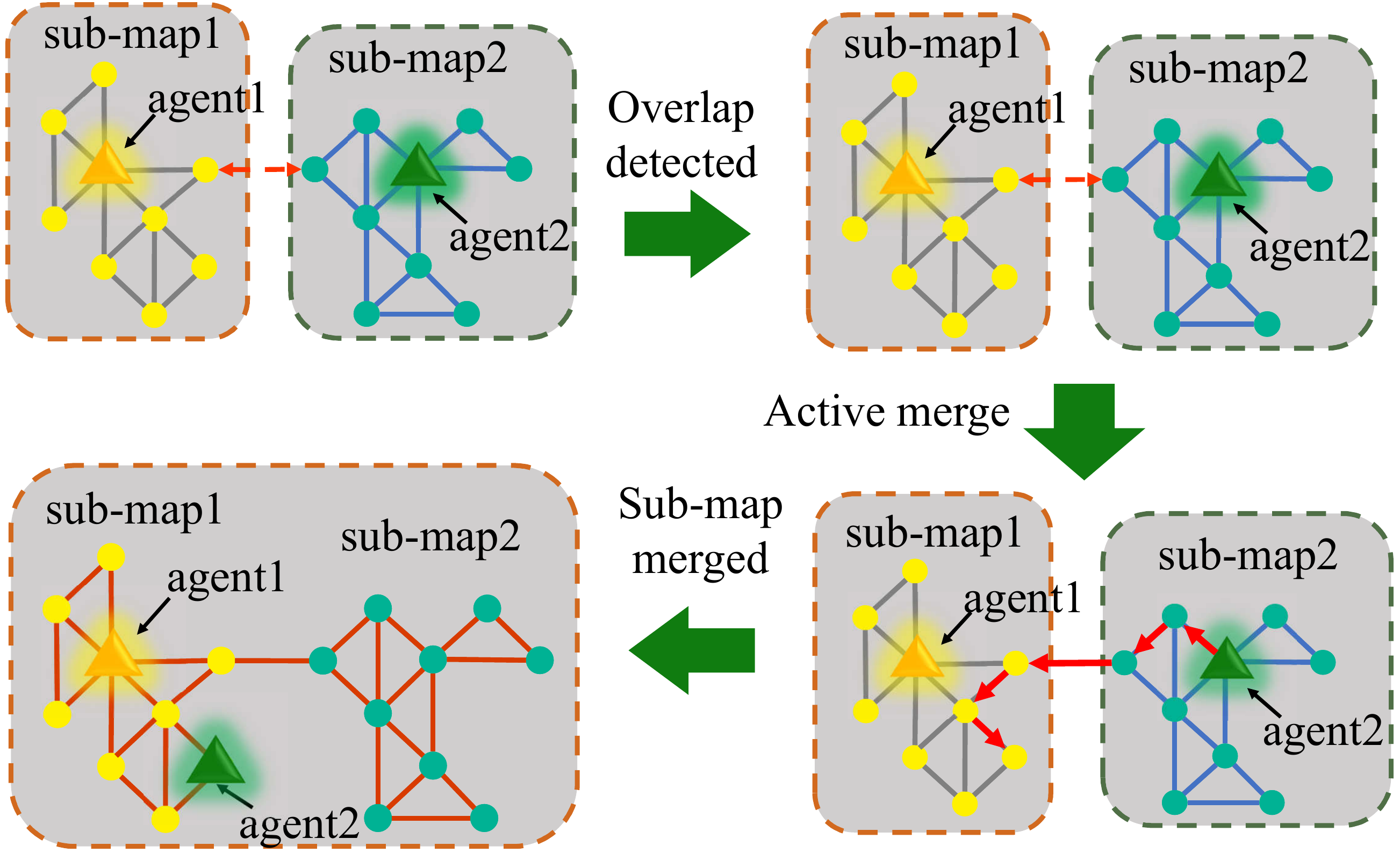}
        \caption{\textbf{Adaptive sub-map merge.}
        A factor graph is used to maintain the inner connections between segments.
        Based on the factor graph, the segments with strong inner connections will be clustered into sub-maps.
        During the exploration process, once the new inner connection is detected, the agent with a shorter distance to the overlap will be used for adaptive merge. 
        This agent will verify and increase the inner connection by actively exploring the potential overlap region.
        This merge process will finish until the inner connection was proved to be an incorrect data association or the sub-maps got merged.}
    \label{fig:mapMerge}
    \end{figure}

\subsection{System Overview}\label{sec:system_overview}
    As shown in Fig.~\ref{fig:OverallFramework}, the entire system consists of multiple agents and an MUI-TARE server. During the exploration, the point clouds collected by the agents are sent to the server, and the server plans and sends the paths to the agents for execution.
    MUI-TARE (Fig.~\ref{fig:OverallFramework}) consists of two major components: an adaptive and active map merging module, and a multi-agent sub-map exploration module. Here, the sub-map means the map that is formed by a group of agents whose relative transform is known.
    The map merging module merges the segments (i.e., the accumulated point cloud) of agents into sub-maps and estimates the relative pose of agents based on the merged sub-maps. Given the point clouds from different agents, our approach detects the potential overlap between the sub-maps via data association: if two series of frames (i.e. the extracted feature at a certain pose) are matched, we say a new inner connection is established between agents. 
    Then, the server plans a path for the nearby agent to verify if this inner connection is a real overlapped area. Once the connection is confirmed, the nearest agent is further navigated to increase the overlap until this new inner connection satisfies the sub-map merge condition. Finally, the overlapped sub-maps are merged into a single sub-map, and the relative position of the agents in this sub-map can be computed.
    
    Based on the sub-maps computed by the map merging module, multi-agent exploration is performed, which uses a hierarchical path planning approach for each sub-map. Specifically, a global coarse path is first planned for agents to visit all the sub-space to be explored. 
    In each sub-space of an agent, a detailed local path is planned to cover all the surfaces in it. 
    MUI-TARE runs the map merging module and the multi-agent exploration module parallel to cover the whole working space.

\subsection{Adaptive Map Merging}\label{sec:adaptive_merge}

    During the exploration process, we maintain a factor graph $G = \{V, E\}$ to represent the inner connections between the segments of agents. Let $V = \{v_{1}, v_{2}, ..., v_{n}\}$ denote the nodes of the graph which represent the segment obtained by each robot. Then, we define edge $E = \{\omega_{i, j}\}$ as the connections between segments, where $\omega_{i, j}$ is the inner connection of node $v_{i}$ and node $v_{j}$. For the purpose of place recognition, the inner connection of two segments is determined by the overlap length (i.e., the number of consecutive matched frames) and place recognition descriptors. Thus, the inner-connection $\omega_{i, j}$ is defined as,
    
    \begin{equation}
    \label{eq:inner-connection}
        {\omega_{i, j}} = \left\{
        \begin{aligned}
            exp^{-\frac{\| {F_i-F_j} \|_{2}^{2}+C_w}{2L_{i, j}^{2} + \epsilon}}, \ i \neq j, \\
            0, \ i = j,
        \end{aligned} 
        \right.
    \end{equation}
    where $F_i, F_j$ indicates the feature descriptors extracted from the overlapped area of $v_i, v_j$, $L_{i, j}$ represents the overlap length between $v_i, v_j$, and $C_w, \epsilon$ are hyper-parameters introduced in AutoMerge \cite{yin2022automerge}.

    \begin{algorithm}[t]
    \caption{Pseudo-code Adaptive overlap estimation}
    \label{algorithm:mergeDetection}
    \begin{algorithmic} [1]
    \While{\textit{ExplorationNotFinished()}}
    \State{\textit{MatchedSequence} $\leftarrow$ \textit{AutoMerge()}}
    \For{\textit{sequence} \textbf{in} \textit{MatchedSequence}}
    \State{\textbf{if} !\textit{IsMatchValid(sequence)} \textbf{then}}
    \State{\indent \textit{Remove(sequence)}}
    \State{\indent \textbf{continue}}
    \State{\textit{inner-connection} $\leftarrow$ \textit{ConnectionEval(sequence)}}
    \State{\textit{T} $\leftarrow$ \textit{TransformEval(sequence)}}
    \State{\textbf{if} \textit{inner-connection \textless \; threshold} \textbf{then}}
    \State{\indent \textit{dist} $\leftarrow$ \textit{EstimateDist(inner-connection)}}
    \State{\indent \textit{ActiveMerge(T, dist)}}
    \State{\textbf{else}}
    \State{\indent \textit{SubmapMerge(T)}}
    \EndFor
    \EndWhile
    \end{algorithmic}
    \end{algorithm}

    \begin{algorithm}
    \caption{Pseudo-code ActiveMerge}
    \label{algorithm:activemerge}
    \textbf{Input:} \textit{T, dist}
    \begin{algorithmic}[1]
    \State{\textit{AgentStats} $\leftarrow$ \textit{GetAgentStatus()}}
    \State{\textbf{if} \textit{AgentStatus} \textbf{is} \textit{BackToOverlap} \textbf{then}}
    \State{\indent \textit{Path} $\leftarrow$ \textit{GetClosestPathToOverlap()}}
    \State{\textbf{else if} \textit{AgentStatus} \textbf{is} \textit{ActiveMerge} \textbf{then}}
    \State{\indent \textit{$G_2$} $\leftarrow$ \textit{PosegraphTransform($G_2$, T)}}
    \State{\indent \textit{Path} $\leftarrow$ \textit{GetLookAheadPath($G_2$, dist)}}
    \State{\indent \textit{VerificationGain} $\leftarrow$ \textit{EvaluateExploreGain()}}
    \State{\indent \textbf{if} \textit{VerificationGain \textless \; threshold} \textbf{then}}
    \State{\indent \indent \textit{PublishInvalidOverlap()}}
    \end{algorithmic}
    \end{algorithm}

    In the merging procedure, only the segments with inner connections larger than the threshold will be clustered and merged into the same sub-map. 
    The relative pose between agents, which is represented by a transform matrix, can be determined by the transform estimated in the map merge.
    However, such stable inner connections usually required long overlap distances which is hard to satisfy when agents are exploring without knowing others' poses.
    Thus, using a shorter sequence of matched frames for potential overlap detection will be helpful to increase the chance an overlap can be detected.
    However, using a shorter sequence of matched frames may cause incorrect data association and large transform matrix errors.
 
    Thus, the adaptive merge will plan a new path for the agent to verify the data association and enhance the inner connection(Fig. \ref{fig:mapMerge}). 
    As shown in Algorithm. \ref{algorithm:mergeDetection}, when a potential overlap was detected between agent $i$ in the sub-map $m_1$ and agent $j $ in the sub-map $m_2$, the inner connection is evaluated and added to the graph $G$. 
    Then, a rough transform matrix $T$ can be estimated by doing the Iterative closest point (ICP) between associated point clouds.
    Based on eq.~\ref{eq:inner-connection}, the value of the inner connection is positively related to the overlap length and negatively related to the feature difference of the overlap. 
    Thus, if the inner connection does not satisfy the merge condition, the planner will choose one agent $i$, whose distance to overlap is shorter, and execute the active merge to increase the inner connection. 
    To help the path planning of active merge, an estimation of the distance that the agent needs to travel is computed:
     \begin{equation}
        \begin{aligned}
        dist = C_w(\frac{1}{\ln{\omega^{t}_{i, j}}}-\frac{1}{\ln{\omega_{thresh}}} )
        \end{aligned}
    \end{equation}
    where the $\omega^{t}_{i, j}$ is the current inner connection, and $\omega_{thresh}$ is the threshold of the inner connection to guide map merging.
    
    As shown in Algorithm. \ref{algorithm:activemerge}, the planner will first plan a path for agent $i$ to move back to the overlapped region by doing $A^*$ search on the past trajectory of agent $i$.
    Once the agent $i$ reaches the overlap, it will be navigated to visit the key pose of agent $j$ outside of the overlap.
    Denote the key pose position of agent $i$ in the sub-map $m_1$ as $\{^{m_1}x_{i}^{0}, ..., ^{m_1}x_{i}^{t}\}$, and denote the key pose position of agent $j$ in the sub-map $m_2$ as $\{^{m_2}x_{j}^{0}, ..., ^{m_2}x_{j}^{t}\}$.
    The trajectory of agent $j$ can be transformed to $\{^{m_1}\widehat{x}_{j}^{0}, ..., ^{m_1}\widehat{x}_{j}^{t}\}$ using $T$.
    Here, we use the greedy strategy to navigate the agent to the closest transformed key pose.
    
    In order to measure the newly obtained overlap during the active merge, we define verification gain as:
     
    \begin{equation}
        \begin{aligned}
            G(i, j, t) = \frac{\omega_{i, j}^t - \omega_{i, j}^{t_0}+C_{\epsilon}}{\|t - {t_0}\|+\epsilon}
        \end{aligned}
    \end{equation}
    where, $\omega^{t}_{i, j}$ is the current value of inner connection, $\omega^{t_0}_{i, j}$ is the value of inner connection when agent start to increase overlap, $\epsilon$ and $C_{\epsilon}$ are constant value.

    During the merge process, if the verification gain is smaller than the threshold, this new connection is regarded as an incorrect connection and gets eliminated from graph $G$. Otherwise, the agent keeps active merge until it meets the sub-map merge requirement. Once the reliable inner connection between sub-maps is established, the sub-maps are merged into a single sub-map as described in \cite{yin2022automerge}.

\vspace{2mm}
\noindent\textbf{Remark.} This module uses the same feature extraction and data association function as the existing AutoMerge paper \cite{yin2022automerge}. However, in order to actively increase the chance that an overlap gets detected, we design the adaptive merge method as shown in lines 3-13 in the Algorithm. \ref{algorithm:mergeDetection} and Algorithm. \ref{algorithm:activemerge}.

\subsection{Multi-agent exploration in sub-maps}\label{sec:explore_submap}
    \begin{figure}[!t]
    	\centering
        \includegraphics[width=\linewidth]{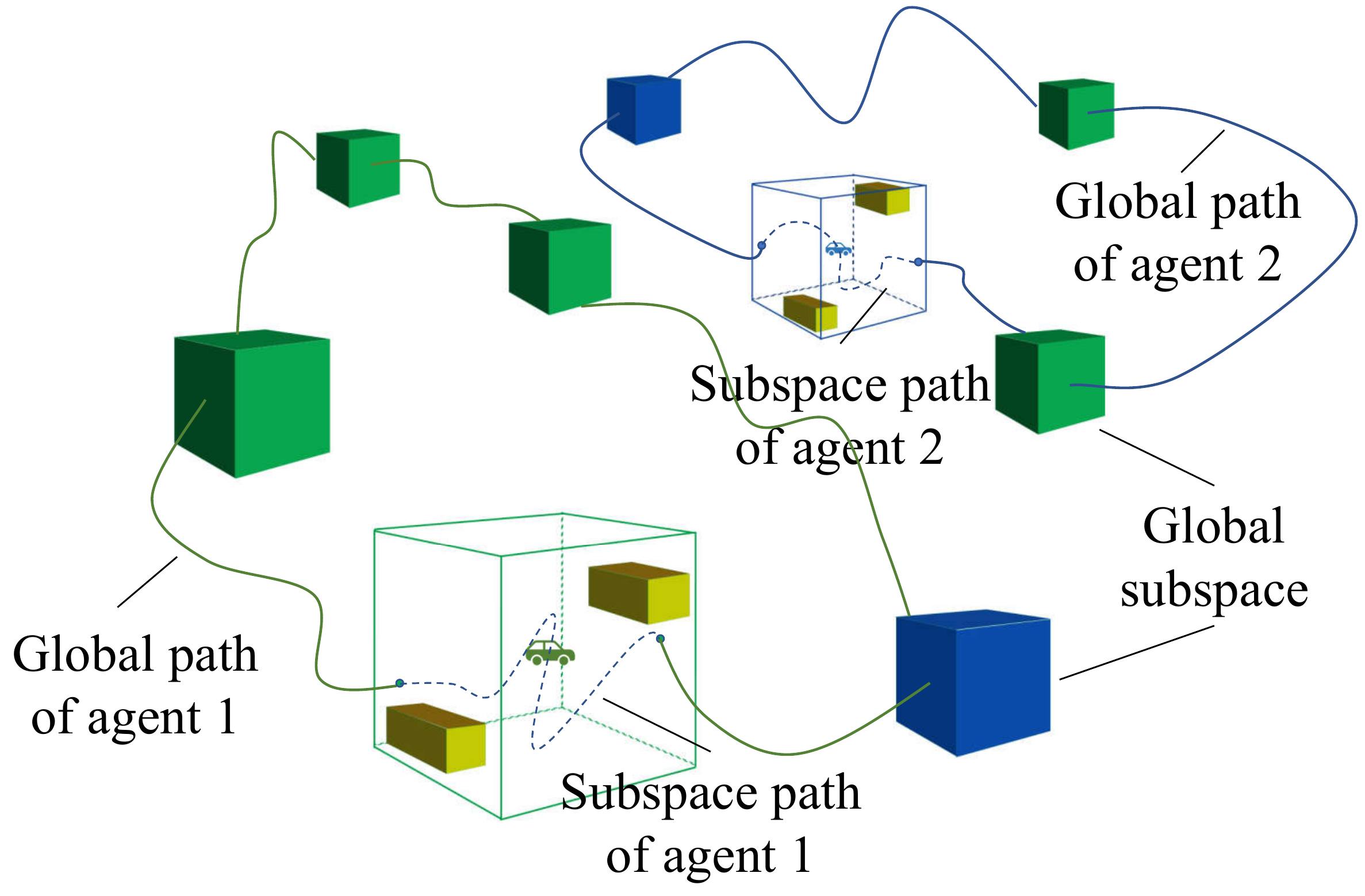}
    	\caption{\textbf{Multi-agent global planning on sub-map.}
            The global planner tries to get the paths that visit all the exploring subspaces in the working space with the minimized longest travel distance by solving the ``min-max multi-depot vehicle routing problem''. 
            Then, a global coarse path is generated for each agent to visit all the sub-space that is assigned to it.
            Here, The green cell represents the sub-space originally explored by agent 1, while the blue cell represents the cell explored by agent 2. Given the merged map and relative pose, a set of global paths is planned that coordinate the agents for exploration.}
    	\label{fig:global-planning-idea}
    \end{figure}

    \begin{algorithm}[t]
    \caption{Pseudocode for hierarchical planning}
    \label{algorithm:global-planning}
    \begin{algorithmic} [1]
    \For{\textit{m := 1 to M}}
        \State{\textit{D'} $\leftarrow$ \textit{GetDistanceMatrix()}}
        \State{$\{\tau_{1}^{global},... \tau_{k}^{global}\} \leftarrow GlobalPlanningSolver(\textit{D'})$} 
        \For{\textit{k := 1 to K}}
            \State{$\tau_{j}^{local} \leftarrow$ \textit{LocalPlanner()}}
            \State{$\tau_{j} \leftarrow$ \textit{PathConcatenate($\tau_{j}^{global}, \tau_{j}^{local})$}}
        \EndFor
    \EndFor
    \Return{$\{\tau_{1},\tau_{2},\dots, \tau_{N}\}$}
    \end{algorithmic}
    \end{algorithm}

    In this work, we extend a state-of-the-art single-agent exploration method TARE to multiple agents.
    As discussed in the previous part, since the transform of agents in each sub-map is given by the adaptive sub-map merge module, we plan the paths for agents based on these sub-maps.
    The total number of the sub-maps is denoted as $M$, and let $m$ be the $m-th$ sub-map, where $m = 1, 2, \dots, M$. Initially, as each sub-map only contains one agent, $M$ equals $N$. As sub-maps get merged, $M$ decreases. Let $K_m$ denote the number of agents in sub-map $m$.
    In each sub-map, a similar sub-spaces-based representation as mentioned in TARE \cite{cao2021tare} is used. It divided the workspace ${Q}$ into even subspaces. 
    Based on the coverage of the subspace, the status of the subspace will be set to ``exploring'', ``explored'', or ``unexplored''. If all the surfaces in the subspace have been covered, the subspace will be labeled as ``explored''. 
    When the subspace contains both uncovered surface and covered surface, the subspace will be assigned an ``exploring'' label. At the same time, for the subspace that is fully uncovered, the label of the subspace will be ``unexplored''.
    
    The path is planned by combining the detailed local path from the subspace planner and the coarse global path from the global planner. 
    For the subspace planners, this paper used the same method as the local planner used in TARE. the local planner samples a set of viewpoints that can cover the surface in the subspace. Then, a local detailed path is planned to travel through all the viewpoints with the shortest distance while satisfying smoothness constraints.
    For the global planner, instead of using a greedy algorithm, this planner optimized the overall path of all agents as shown in Fig. \ref{fig:global-planning-idea}.

    The goal of the global planner is planning a set of global path $\tau_{global} = \{\tau_{1}^{global},... \tau_{K_m}^{global}\}$ for $K_m$ agents in the sub-map $m$ to travel through all the ``exploring'' subspace. 
    Since the agents travel in parallel, in order to minimize the overall travel time, the planner tries to minimize the maximum travel distance among all the agents:
    \begin{equation}
        \begin{aligned}
            &\min (dist(\tau^{global}))=\\
            &\min (\max(dist(\tau_{1}^{global}), ..., dist(\tau_{K_m}^{global})))
        \end{aligned}
    \end{equation}
    Thus, this problem can be formulated as a variant of the standard multi-depot vehicle routing problem (MDVRP) as ``min-max MDVRP''. Since this problem is NP-hard, heuristic methods are used to solve it \cite{carlsson2009solving}. 

    As shown in the Algorithm \ref{algorithm:global-planning}, for $M$ sub-maps in the current working space, the distance between all the exploring cells and agents in each sub-map will be calculated to form the distance matrix $D'$. This distance is obtained by doing the $A^*$ search on the roadmap acquired during the exploration process. Then, we use the min-max MDVRP solver to solve the minimum longest path using the distance matrix $D'$. Given this solution, $K_m$ global coarse paths $\{\tau_{1}^{global},... \tau_{K_m}^{global}\}$ can be formed for $K_m$ agents in the sub-map $m$. 
    With the local path planned by the local planner, each agent replaces the part of the global path that falls in the subspace with its own local path to create the full exploration path.
    Finally, these full exploration paths $\{\tau_{1},\tau_{2},\dots, \tau_{N}\}$ are sent to all the $N$ agents for execution.
    
\vspace{2mm}
\noindent\textbf{Remark.} 
This multi-agent exploration uses a similar global representation and local planning method as in TARE \cite{cao2021tare}.
Additionally, we design a new cooperative global planner for multi-agent global planning  as shown in the Algorithm \ref{algorithm:global-planning}.

\section{Experiments}
\label{sec:experiments}

This section presents experimental results and discussions of MUI-TARE.
We compare it with several exploration methods \cite{yu2021smmr, 8967932, cao2021tare} and show the advantage of our MUI-TARE.

\begin{figure*}[t]
    \centering
    \includegraphics[width=\linewidth]{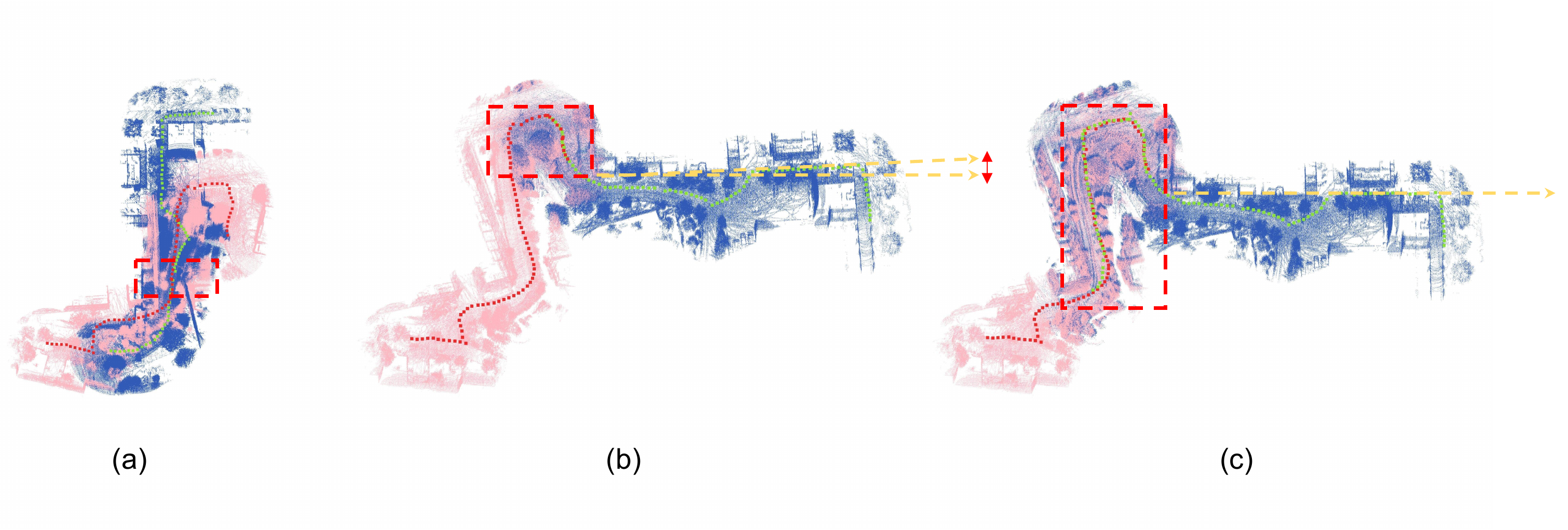}
    \caption{\textbf{The trajectory and point clouds of different overlap.}
    Green dots and red dots in the trajectory represent the selected frames of two agents. In (a), the segments get merged with the wrong data association. In (b), the overlap is correct, however, the deviation is still large as shown by the red arrow. In (c), the map is better merged  by increasing the overlap using adaptive merge.}
    \label{fig:activemerge_result}
\end{figure*}

\begin{figure}[ht]
\centering
    \includegraphics[width=\linewidth]{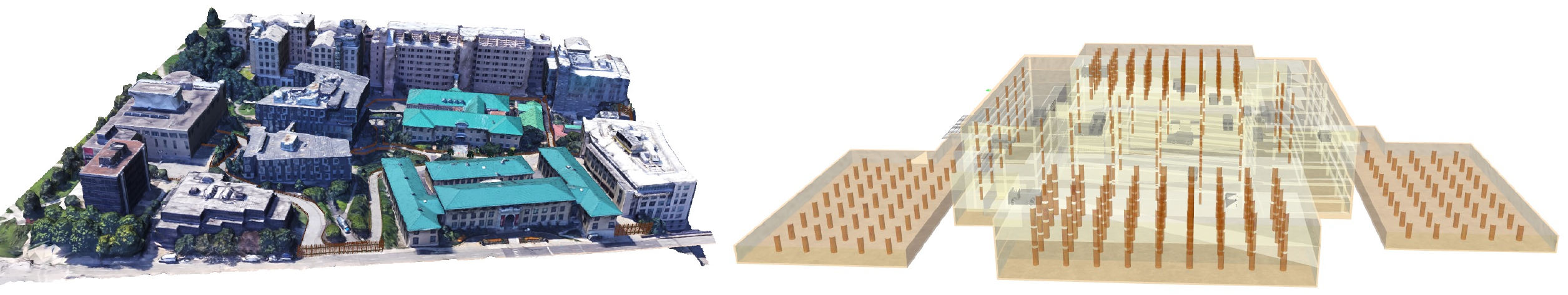}
    \caption{\textbf{Simulation environments.}
    The left side is the campus environment, while the right side is the garage environment.}
    \label{fig:experiment_env}
\end{figure}

\subsection{Experiment Setup} \label{subsec:exp_setup} 
We test MUI-TARE in several different simulation environments.
The simulation experiment consists of multiple robots and a central server, each on a different computer.
The robots are simulated in Gazebo, and each robot is equipped with a simulated 360-degree Velodyne Lidar. The collected point clouds are used for map merging and path planning for exploration. In addition, an IMU is fused with the Lidar data for state estimation. During the experiment, the maximum speed of the robot is 3 m/s.
In our implementation, the simulation environment is divided evenly into sub-spaces, where each sub-space is a 10m×10m×5m cell. 
We use the ROS multi-master communication package \cite{tiderko2016ros} for data exchange between the agents and the server.
As we only transfer the point clouds collected by agents, and the waypoints computed by the server. The average bandwidth needed for a single agent is 786 KB/s in our experiments, which is similar to the recent work~\cite{yu2021smmr}.
As shown in Fig. \ref{fig:experiment_env}, the experiments are done in two simulation environments from \cite{cao2022autonomous}.

\begin{itemize}
    \item  Campus (340m $\times$ 340m): A part of the Carnegie Mellon University campus, containing undulating terrains and convoluted environment layout.
    \item Multi-storage Garage (140m $\times$ 130m, 5 floors): A complex 3D environment with multiple floors and sloped terrains.
\end{itemize}

\begin{figure}[t]
    \centering
    \includegraphics[width=\linewidth]{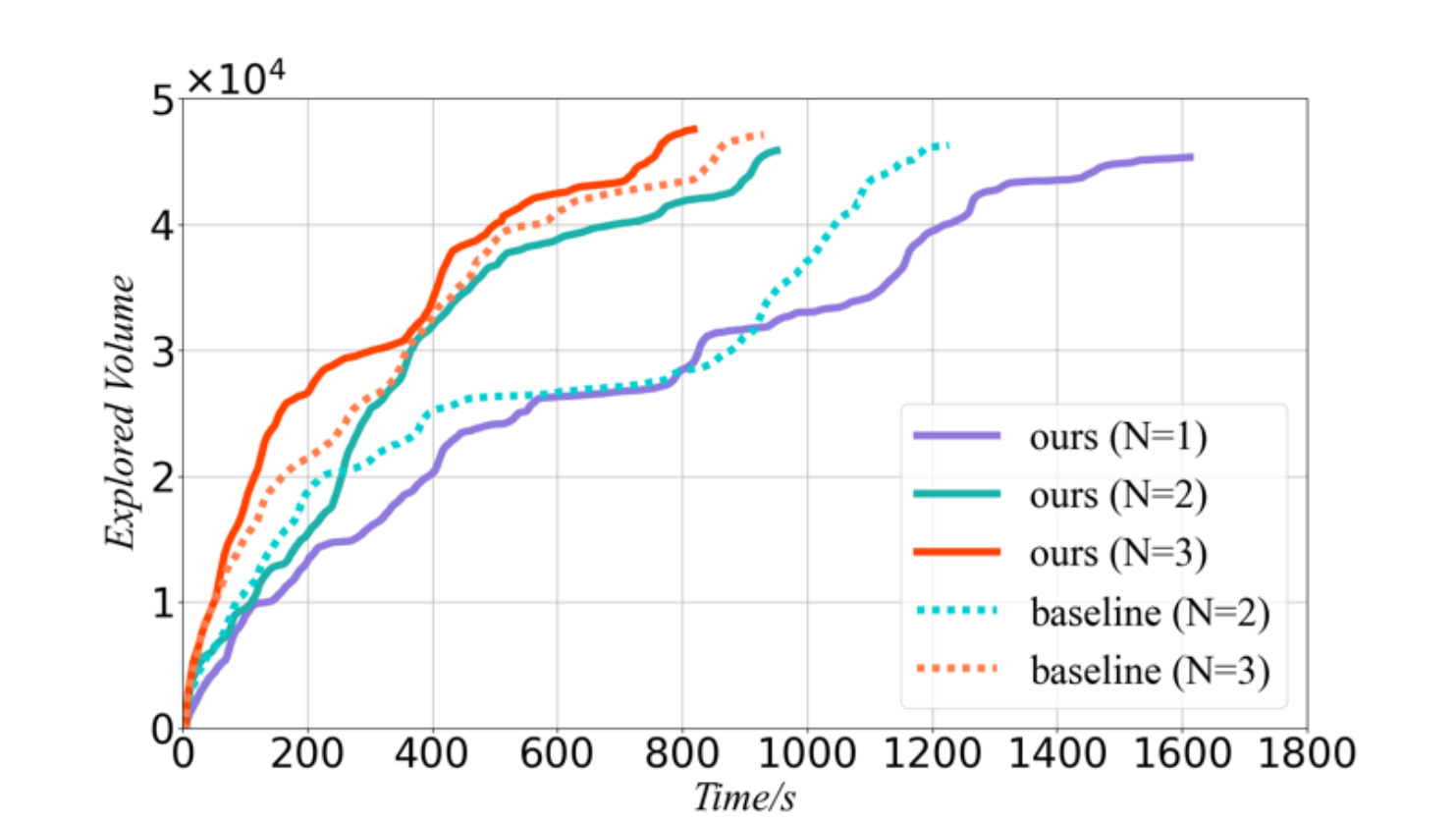}
    \caption{\textbf{Result of Test 1.}
    Comparison of our adaptive merge with the TRUST-Explorer in terms of exploration volume versus time.}
    \label{fig:activemerge_efficiency}
\end{figure}

\begin{figure*}[t]
    \centering    \includegraphics[width=\linewidth]{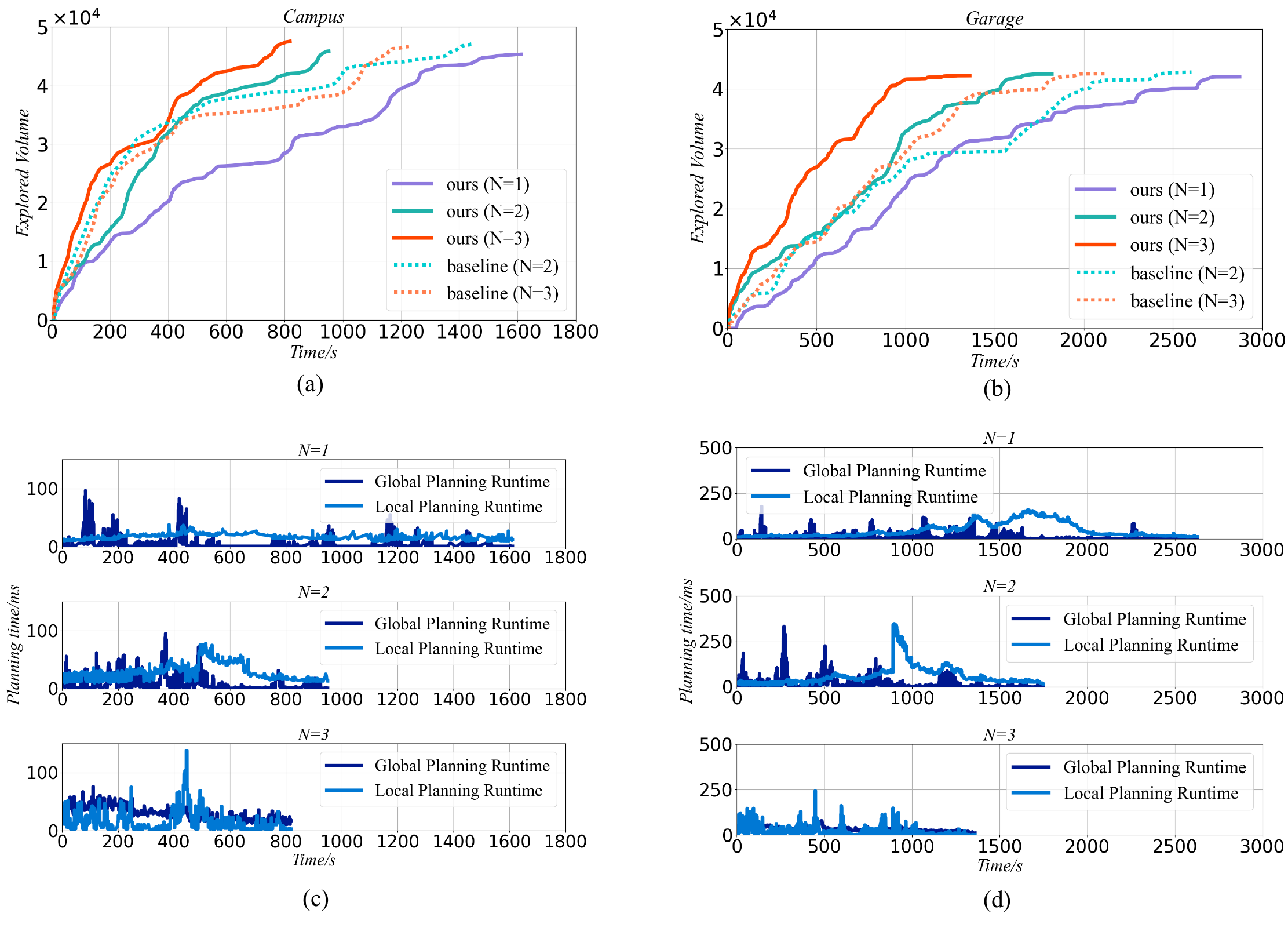}
    \caption{\textbf{The exploration efficiency \& planning runtime in both environments.}
    The left side is the result of the campus environment, while the right side is the result of the garage. As shown in the plot, our method shows significant improvement over the mTARE, while keeping the planning runtime in a reasonable range.}
\label{fig:multi_explore_efficiency}
\end{figure*}

\subsection{Adaptive Merge Evaluation} \label{subsec:AM_results}

\subsubsection{Baselines}
To show the performance of our Adaptive Merge algorithm, we compare MUI-TARE against the following baseline methods. These baselines are designed by adapting the ideas in the respective paper to our test scenarios.
\begin{itemize}
    \item \textbf{TRUST-Explorer}: leverages the idea in \cite{8967932}. It verifies the overlap between two agents every time when the potential overlap is detected. The verification is done by navigating one robot to repeat an excessive amount of the other agent's trajectory. However, this method is designed mainly for a 2D environment. We adapt this idea in our framework by letting one agent repeat a fixed amount of the other agent's trajectory.
    \item \textbf{SMMR-Explorer}: leverages the idea in~\cite{yu2021smmr}. It uses an aggressive merge strategy where the sub-maps are merged once the overlap is detected without verification. 
\end{itemize}

\begin{table}[t]
\tabcolsep=0.1cm
\renewcommand{\arraystretch}{1.6}
\centering
\caption{Exploration efficiency for Adaptive Merge compare with TRUST-Explorer.}
\begin{tabular}{ccccc}
    \hline
        \multirow{2}{*}{\bfseries Metrics} & 
        \multicolumn{2}{c}{\bfseries TRUST-Explorer} & 
        \multicolumn{2}{c}{\bfseries ours} \\
        & N=2 & N=3 & N=2 & N=3 \\
    \hline
    Exploration Time (s)  & 1226.89 & 930.33 & 951.45 & 818.36 \\
    \hline
\end{tabular}
\label{table:activemerge_table}
\end{table}

\subsubsection{Results and Discussion}

In comparison with TRUST-Explorer, Fig. \ref{fig:activemerge_efficiency} shows the explored volume against time while Table \ref{table:activemerge_table} shows the exploration efficiency. We observed that our method is 29\% faster for 2 agents, and 14\% faster for 3 agents.

We then compare our method against SMMR-Explorer.
As shown in Fig.~\ref{fig:activemerge_result} (a), the aggressive strategy in SMMR-Explorer can lead to incorrect data association, which results in an incorrect map after merging. As shown in Fig. \ref{fig:activemerge_result} (b), SMMR-Explorer can also lead to large errors in the computed transform matrix. In comparison, Fig. \ref{fig:activemerge_result} (c) shows the merged map using our approach which results in robust map merging and small errors in the transform matrix.
To summarize, due to the adaptive exploration strategy, our MUI-TARE achieves higher exploration efficiency than TRUST-Explorer and more robust map merging than SMMR-Explorer.


\subsection{Multi-Agent Exploration Evaluation} \label{subsec:MX_results}
\begin{table}[t]
\tabcolsep=0.1cm
\renewcommand{\arraystretch}{1.6}
\centering
\caption{Exploration efficiency \& runtime for multi-agent exploration at campus and garage compare with mTARE.}
\begin{tabular}{c|ccccc}
    \hline
        \multirow{2}{*}{\bfseries Env} &
        \multirow{2}{*}{\bfseries Metrics} & 
        \multicolumn{2}{c}{\bfseries mTARE} & 
        \multicolumn{2}{c}{\bfseries ours} \\[2pt]
        && N=2 & N=3 & N=2 & N=3 \\
    \hline
    \multirow{3}{*}{\rotatebox{90}{Campus}}
        &Exploration Time(s)  & 1450.35 & 1234.36 & 951.45 & 818.36 \\
        &Local Plan Runtime (ms) & 16.40 & 22.23 & 28.72 & 29.60 \\
        &Global Plan Runtime (ms) & 8.18 & 8.34 & 11.48 & 17.30 \\
    \hline
    \multirow{3}{*}{\rotatebox{90}{Garage}}
        &Exploration Time (s)  & 2502.13 & 1989.09 & 1665.11 & 1022.92 \\
        &Local Plan Runtime (ms) & 63.68 & 38.78 & 33.20 & 45.77\\
        &Global Plan Runtime (ms) & 9.99 & 9.99 & 18.93 & 23.19\\
    \hline
\end{tabular}
\label{table:explore_efficiency}
\end{table}

This section compares our MUI-TARE with a naive extension of TARE to multiple agents. Specifically, we introduce multi-agent TARE (mTARE) as a baseline, which plans for each agent independently without planning multiple agents together after merging.

\subsubsection{Results}
We compare our method against mTARE in the garage environment with 2 and 3 agents.
As shown in Fig. \ref{fig:multi_explore_efficiency}, the upper plots show the explored volume versus exploration duration while the lower plots show the runtime of the planning during the exploration.
We present the exploration time and planning time in Table \ref{table:explore_efficiency}.
It shows that the exploration time of our method is 52\% faster than mTARE for 2 agents.
For 3 agents, our MUI-TARE is 51\% faster than mTARE.
This result verifies the benefit of the multi-agent global planning strategy used in MUI-TARE. Using the global path planned from it, our agents will be able to reduce the cost.

Additionally, Table \ref{table:explore_efficiency} shows the average local planning and global planning runtime per call in MUI-TARE and mTARE. 
We observed that, the planning time has no obvious increment as the number of agents increases.
Finally, as shown in Fig. \ref{fig:multi_explore_efficiency} (a), by increasing the number of agents from one to two, the time required to finish the exploration (i.e., the time when the explored volume stop increasing) is reduced by roughly 64\%.
However, by increasing the number of agents from two to three, this reduction in exploration time is only about 16\%.
This indicates a diminishing return when increasing the number of agents in a exploration task.

\section{Conclusions and Future Work}
\label{sec:future-work}
This paper presents MUI-TARE, a multi-agent cooperative method for exploration with the unknown initial position.
MUI-TARE intelligently balances the robustness of sub-map merging and exploration efficiency, by using an adaptive approach for map merging.
Additionally, MUI-TARE extends the recent single-agent hierarchical exploration strategy to multiple agents in a cooperative manner by planning path of multiple agents together after their sub-maps are merged to further improve exploration efficiency.
Our numerical results verifies the benefits of our approach.

For future work, one can consider extending MUI-TARE to a decentralized or distributed version where no global communication is required. 
In addition, this work mainly focuses on the problem where the initial positions of agents are completely unknown. One can extend this method to address the case where partial prior knowledge about the initial agent position is available.

\bibliographystyle{IEEEtran}
\bibliography{IEEEexample}
\end{document}